\documentclass[acmlarge]{acmart}

\usepackage{soul}
\usepackage{multirow}
\usepackage{float}                  
\usepackage{subfig}                 
\usepackage{color}

\definecolor{pink}{RGB}{255,0,255}

\AtBeginDocument{%
  }

\setcopyright{acmcopyright}
\copyrightyear{2018}
\acmYear{2018}
\acmDOI{XXXXXXX.XXXXXXX}

\begin{document}

\title{You Only Forward Once: Prediction and Rationalization in A Single Forward Pass}

\author{Han Jiang}
\email{jh-better@csu.edu.cn}
\affiliation{%
  \institution{School of Computer Science and Engineering, Central South University}
  \city{Changsha}
  \state{Hunan}
  \country{China}
}
\author{Junwen Duan}
\authornote{corresponding author}
\email{jwduan@csu.edu.cn}
\affiliation{%
  \institution{School of Computer Science and Engineering, Central South University}
  \city{Changsha}
  \state{Hunan}
  \country{China}
}

\author{Zhe Qu}
\email{zhe_qu@csu.edu.cn}
\affiliation{%
  \institution{School of Computer Science and Engineering, Central South University}
  \city{Changsha}
  \state{Hunan}
  \country{China}
}

\author{Jianxin Wang}
\email{jxwang@mail.csu.edu.cn}
\affiliation{%
  \institution{School of Computer Science and Engineering, Central South University}
  \city{Changsha}
  \state{Hunan}
  \country{China}
}






\renewcommand{\shortauthors}{Jiang et al.}

\begin{abstract}
Unsupervised rationale extraction aims to extract concise and contiguous text snippets to support model predictions without any annotated rationale. 
Previous studies have used a two-phase framework known as the Rationalizing Neural Prediction (RNP) framework, which follows a generate-then-predict paradigm. 
They assumed that the extracted explanation, called rationale, should be sufficient to predict the golden label.
However, the assumption above deviates from the original definition and is too strict to perform well.
Furthermore, these two-phase models suffer from the interlocking problem and spurious correlations.
To solve the above problems, we propose a novel single-phase framework called You Only Forward Once (YOFO), derived from a relaxed version of rationale where rationales aim to support model predictions rather than make predictions.
In our framework, A pre-trained language model like BERT is deployed to simultaneously perform prediction and rationalization with less impact from interlocking or spurious correlations. 
Directly choosing the important tokens in an unsupervised manner is intractable.
Instead of directly choosing the important tokens, YOFO gradually removes unimportant tokens during forward propagation.
Through experiments on the BeerAdvocate and Hotel Review datasets, we demonstrate that our model is able to extract rationales and make predictions more accurately compared to RNP-based models.
We observe an improvement of up to 18.4\% in token-level F1 compared to previous state-of-the-art methods. We also conducted analyses and experiments to explore the extracted rationales and token decay strategies. 
The results show that YOFO can extract precise and important rationales while removing unimportant tokens in the middle part of the model.
\end{abstract}

\ccsdesc[500]{Computing methodologies~Natural language processing}
\ccsdesc[500]{General and reference~Reliability}
\ccsdesc[500]{Computer systems organization~Reliability}

\keywords{Rationale Extraction, Pretrained Language Model, Token Reduction}

\received{20 February 2007}
\received[revised]{12 March 2009}
\received[accepted]{5 June 2009}

\maketitle

\section{Introduction}
Deep learning has achieved remarkable success in the field of natural language processing (NLP)~\cite{YoonKim2014ConvolutionalNN, AshishVaswani2017AttentionIA, AlexWang2018GLUEAM, AlexWang2019SuperGLUEAS}, especially with the deployment of pre-trained language models like BERT~\cite{JacobDevlin2018BERTPO}. 
Despite their impressive performance on various tasks, deep learning AI systems have long faced the challenge of interpretability~\cite{sun2021interpreting}.
This lack of interpretability is particularly problematic in domains such as medicine and finance, where understanding the reasoning behind AI decisions is crucial.

Extractive rationalization~\cite{lei-etal-2016-rationalizing, bastings-etal-2019-interpretable, FR} is an approach to open the black-box AI system by extracting text snippets that support the final model predictions. 
Some researches~\cite{deyoung2019eraser, li2022unifying} have demonstrated the effectiveness of supervised rationale extraction. 
However, obtaining annotations for rationales in real-world scenarios can be challenging. 
In this paper, we will concentrate on unsupervised rationale extraction, where no rationale annotation is available during the training process.
\begin{figure}
	\centering
	\subfloat[RNP]{\includegraphics[width=.27\columnwidth]{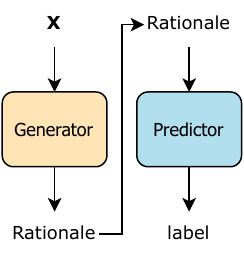}\label{fig:RNP}}
	\subfloat[FR]{\includegraphics[width=.3\columnwidth]{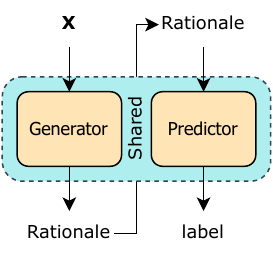}\label{fig:FR}} 
	\subfloat[YOFO(\textit{ours})]{\includegraphics[width=.3\columnwidth]{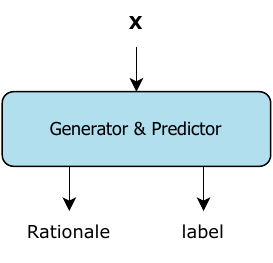}\label{fig:YOFO}}
	\caption{Comparison of our methods (YOFO) with previous typical works in RNP framework. a). the basic frameworks proposed in~\cite{lei-etal-2016-rationalizing}. b). The folded Rationalization model from~\cite{FR} shares the backbone of the generator and predictor. c). Our proposed framework, generates the rationale and prediction with a single forward pass.}
\end{figure}

Most previous works on unsupervised rationale extraction have primarily relied on the Rationalizing Neural Prediction (RNP) framework~\cite{lei-etal-2016-rationalizing}, shown in Figures \ref{fig:RNP} and \ref{fig:FR}. 
It is assumed that rationales should be short, coherent, and provide sufficient information for accurate predictions~\cite{lei-etal-2016-rationalizing}. 
Specifically, the generator extracts rationales, which are then input to the predictor for final predictions. 
The entire model is trained end-to-end using either REINFORCE~\cite{REINFORCE} or reparameterization tricks~\cite{bastings-etal-2019-interpretable, diffmask} to address the non-differentiable problem. 
However, these two-phase models face performance degradation problems like interlocking~\cite{interlocking} and spurious correlations~\cite{INVRAT}. 
Interlocking occurs when the predictor overfits meaningless but distinguishable rationales generated by the still-untrained generator, leading the generator to converge to a sub-optimal model that tends to select these uninformative rationales~\cite{interlocking}. 
The primary problem with spurious correlations arises from the classification process, which relies on meaningless texts highly correlated with labels~\cite{INVRAT}.
Several approaches~\cite{interlocking, FR, MGR, DR, INVRAT, inter-rat} have been proposed to alleviate this problem, but the fundamental issue in such models remains unsolved. 
We believe the fundamental problem that causes performance degradation is the two-phase forward mechanism.
In the two-phase models, prediction is directly decided by extracted rationale, which directly causes the interlocking problem and indirectly enhances the impact of spurious features.
Furthermore, recent experiments~\cite{chen2022can} indicate unknown obstacles that make fine-tuning large pre-trained models on the self-explaining rationalization framework difficult.
This limitation has a negative impact on the overall performance of the model on downstream tasks.

The definition of rationales in RNP~\cite{lei-etal-2016-rationalizing} framework is still overly strict. 
Instead, we propose that a rationale should be: a) short, coherent, and human-readable; b) sufficient to support the model predictions rather than generating them. 
"Supporting the predictions" differs from "making predictions." The former can be framed as a YES/NO problem (i.e., whether it can or cannot support), while the latter is a generative problem (i.e., generating the label). 
This relaxed definition is more reasonable and aligns better with human expectations of the model's capabilities.

Based on these observations, we introduce a new framework called You Only Forward Once (YOFO). 
YOFO, depicted in Figure \ref{fig:YOFO}, is a pre-trained language model that generates predictions and rationales simultaneously in a single forward pass, eliminating the requirement for separate generator and predictor components. 
Additionally, YOFO does not make predictions directly based on extracted rationales so it has less effect by interlocking and spurious correlations.
Instead, it progressively identifies important tokens (i.e., rationales) as the layers deepen, which are then employed to support the model's predictions.
During the experiments, it was realized that directly selecting important tokens in an unsupervised manner is challenging. 
YOFO takes a different approach by gradually removing unimportant tokens as the layers deepen. 

Extensive experiments were conducted on two unsupervised rationale extraction datasets: the BeerAdvocate dataset~\cite{beer} and the Hotel Review dataset~\cite{hotel}. 
The experimental results demonstrate that YOFO outperforms all existing state-of-the-art methods, showcasing the superior performance and effectiveness of our proposed approach. 
Additionally, thorough experiments were conducted to investigate the effectiveness of extracted rationales and different token decay strategies. 
Our contributions can be summarized as follows:
\begin{itemize}
    \item We deviate from the traditional framework of a generate-then-predict paradigm and derive from the relaxed definition of rationale. We introduce a novel framework called You Only Forward Once~(YOFO) where pre-trained language models are leveraged to generate both rationales and predictions in a single forward pass in the unsupervised rationale extraction scenario.
    \item We evaluated our method on the BeerAdvocate and Hotel Review datasets, showcasing its superiority over state-of-the-art approaches. Our method achieved a substantial improvement of up to 18.4\% in token-level F1 for rationale extraction.
    \item We have conducted extensive analyses and experiments to explore the extracted rationales and token decay strategies and found that YOFO can extract precise and important rationales while removing unimportant tokens in the middle part of the model.
\end{itemize}

\section{Related Work}
\subsection{Rationale Extraction}
\label{section:related_rationale}
The rationalization framework, known as RNP~\cite{lei-etal-2016-rationalizing}, is flexible and offers a unique advantage: certification of exclusion. This means that any unselected input is guaranteed to have no contribution to the prediction~\cite{interlocking}. 
However, training this method is difficult. Various approaches have been proposed to improve RNP from different angles to address this challenge.

\paragraph{Gradient Flows} 
The base RNP framework utilizes REINFORCE~\cite{REINFORCE}, but this leads to training instability and poor performance.
To overcome these issues, HardKuma~\cite{bastings-etal-2019-interpretable} introduces re-parameterization tricks and replaces the Bernoulli distribution with the rectified Kumaraswamy distribution, which stabilizes the training process.
In FR~\cite{FR}, the encoder's parameter is shared between the generator and predictor. This ensures that the encoder's gradient is more reasonable because it can see both full texts and rationales.
3Players~\cite{3player} controls the complementary rationale to be meaningless, resulting in more meaningful generated rationales.

\paragraph{Interlocking}
The interlocking problem was initially proposed by A2R~\cite{interlocking}. This problem arises when the generator fails to identify important tokens, leading to sub-optimal rationales and consequently affecting the performance of the generator-predictor system.
Many previous researchers have developed mechanisms to address this issue and provide the predictor with full-text information~\cite{DMR, interlocking, MGR}.
DMR~\cite{DMR} aimed to align the distributions of rationales with the full input text, both in the output space and feature space.
A2R~\cite{interlocking} enhances the predictor's understanding of the full text by introducing a soft rationale.
MGR~\cite{MGR} involves multiple generators with different initializations to allow the predictor to make predictions based on various rationales, simulating the interlocking problem.
DR~\cite{DR} limits the Lipschitz constant of the predictor, making it more robust.

Fundamentally, YOFO does not exhibit the interlocking problem. This is because YOFO deviates from the generate-then-predict paradigm and does not heavily rely on rationales when making predictions.

\paragraph{Spurious correlations}
The spurious correlations problem refers to the tendency of the predictor to make predictions based on meaningless texts that are highly correlated with the labels.
The RNP framework is significantly affected by this issue, as the predictor relies on text that has been previously extracted to make predictions.
Several approaches have been proposed to simulate and address this problem~\cite{INVRAT, inter-rat}.
INVRAT~\cite{INVRAT} introduced an environment-agnostic predictor to identify and recognize spurious correlations.
Inter-RAT~\cite{inter-rat} aimed to eliminate spurious correlations through backdoor adjustment techniques.

Unlike the RNP framework, YOFO relies on the full text to make predictions, gradually generating important texts as rationales. As a result, YOFO is less susceptible to the spurious correlations problem.

\subsection{Token Reduction}
Token reduction aims to accelerate the inference time of pre-trained language models by removing unimportant tokens that do not affect model performance. Previous researchers have proposed various methods for achieving this goal~\cite{Power-BERT, LTP, TR-BERT, guan2022transkimmer}.
Power-BERT~\cite{Power-BERT} utilizes attention weights to determine which tokens can be removed. LTP~\cite{LTP} introduces a soft mask to learn which tokens should be deleted. TR-BERT~\cite{TR-BERT} and Transkimmer~\cite{guan2022transkimmer} select tokens discretely and employ the REINFORCE~\cite{REINFORCE} and reparameterization tricks, respectively, to overcome the non-differentiable problem associated with discrete token removal.

These models are designed to accelerate the inference of pre-trained language models by removing tokens early and extensively, while retaining a few meaningful tokens in the final layer.
Our work aims to enhance the interpretability of the model, utilizing token reduction as a tool for selecting important tokens. YOFO selectively deletes tokens once the model has gathered sufficient information from them, while keeping some insignificant words to form rationales that are easily understandable by humans.

\section{Background}
In this section, we provide a brief introduction to the RNP framework. 
In this framework, the rationale extraction problem is modeled as a generate-then-predict end-to-end pipeline~\cite{lei-etal-2016-rationalizing, bastings-etal-2019-interpretable, FR}.

Formally, let's assume that the input texts $\mathbf{X}$ and their corresponding labels $\mathbf{y}$ are from the training dataset $\mathcal{D}$. 
Unlike a black-box AI system that directly models $P(\mathbf{y}\mid\mathbf{X})$, this framework extracts text snippets $\mathbf{Z}$ from $\mathbf{X}$, where $\mathbf{Z}$ is a subset of $\mathbf{X}$ and can serve as an explanation for the prediction. 
The prediction is then based on both $\mathbf{Z}$ and $\mathbf{X}$. 
Therefore, the problem can be modeled using Equation~(\ref{eq:probs}):
\begin{equation}
    \begin{gathered}
        \label{eq:probs} P(\mathbf{y}\mid\mathbf{X})=P(\mathbf{y}\mid\mathbf{Z}, \mathbf{X})P(\mathbf{Z}\mid\mathbf{X})\\
        =P(\mathbf{y}\mid\mathbf{Z})P(\mathbf{Z}\mid\mathbf{X}).
    \end{gathered}
\end{equation}
the Equation~(\ref{eq:probs}) is simplified by the graph model $\mathbf{y}\leftarrow \mathbf{Z} \leftarrow \mathbf{X}$ in which $\mathbf{X}\perp\!\!\!\perp\mathbf{y}\mid\mathbf{Z}$. 

In the RNP framework, a binary mask $\mathbf{m}\in\{0, 1\}^{L}$ is utilized to calculate $\mathbf{z}$ using the element-wise product $\mathbf{z}=\mathbf{x}\odot\mathbf{m}$. 
Here, $m^i=1$ indicates that the $i$-th token is selected as part of the rationale, and vice versa.
REINFORCE~\cite{REINFORCE} or reparameterization tricks~\cite{bastings-etal-2019-interpretable} are commonly used to overcome the non-differentiable problem.
To further enhance the optimization process, a sparsity and contiguous regularization term are incorporated into the final loss.
The overall procedures can be expressed by the following formulas:
\begin{gather}
    \mathbf{m}=Gen(\mathbf{x};\boldsymbol{\theta}) \\
    \hat{y}=Pred(\mathbf{m}\odot\mathbf{x};\boldsymbol{\omega}) \\
    l=\mathrm{loss\_fn}(\hat{y}, y) + \boldsymbol{\Omega}(\mathbf{m}) \\
    \boldsymbol{\Omega}(\mathbf{m})=\sum_{i=1}^L|m^i|+\sum_{i=1}^{L-1}|m^{i+1}-m^{i}|
\end{gather}
where the first and the second term of ${\boldsymbol{\Omega}}(\cdot)$ is for sparsity and contiguous control, respectively.

These two-phase models often encounter performance degradation issues~\cite{3player, interlocking}. 
Moreover, in this setup, the text needs to go through the encoder twice to sequentially obtain rationales and predictions. 
The integration of pre-trained language models into the RNP framework can be problematic due to the significant latency introduced by pre-trained language models.

\section{You Only Forward Once}
To address these challenges, we propose a novel framework called You Only Forward Once (YOFO) that allows for the simultaneous generation of predictions and rationales.
In this case, we directly model the whole question as $P(\mathbf{y}, \mathbf{Z}\mid\mathbf{X})$, instead of the Equation~(\ref{eq:probs}), to predict and rationalize simultaneously.
YOFO consists of two key components: the \textit{Token Selection} mechanism and the \textit{Contiguous Penalty}. 
The \textit{Token Selection} mechanism selects important tokens during forward propagation to support model predictions, while the \textit{Contiguous Penalty} encourages the generation of more meaningful and human-readable rationales. 
These components synergistically guide the pre-trained language model to achieve accurate prediction and explanation simultaneously. 
Detailed introductions of these components are provided in the following sections.

\subsection{Token Selection}
\label{part:tr}
The token selection mechanism aims to identify crucial tokens that support the model's predictions. 
As the layer goes deeper, the selection process gradually picks out significant words. 
Before reaching the final layer, important text snippets are selected while maintaining a certain level of sparsity.
Determining which tokens should be considered important is a critical strategy. 
However, without rationale annotations, directly selecting significant tokens without impeding information flow between transformer layers poses a challenge.
In this paper, we take the opposite approach.

As an alternative, we aim to gradually eliminate unnecessary tokens as we progress through the layers.
During each layer, unimportant tokens are removed, and the remaining tokens are passed to the next layer. 
Tokens can interact with others due to the multi-head attention mechanism.
We expect that the model can identify and remove the unimportant tokens as the layer goes deeper.
Our framework utilizes Transkimmer~\cite{guan2022transkimmer} as a token erasing method, which employs reparameterization techniques to handle the non-differentiable problem. 
It is important to note that any token-selection methods, such as TR-BERT~\cite{TR-BERT}, can be deployed in our framework.

\begin{figure*}
    \centering
    \includegraphics[scale=0.65]{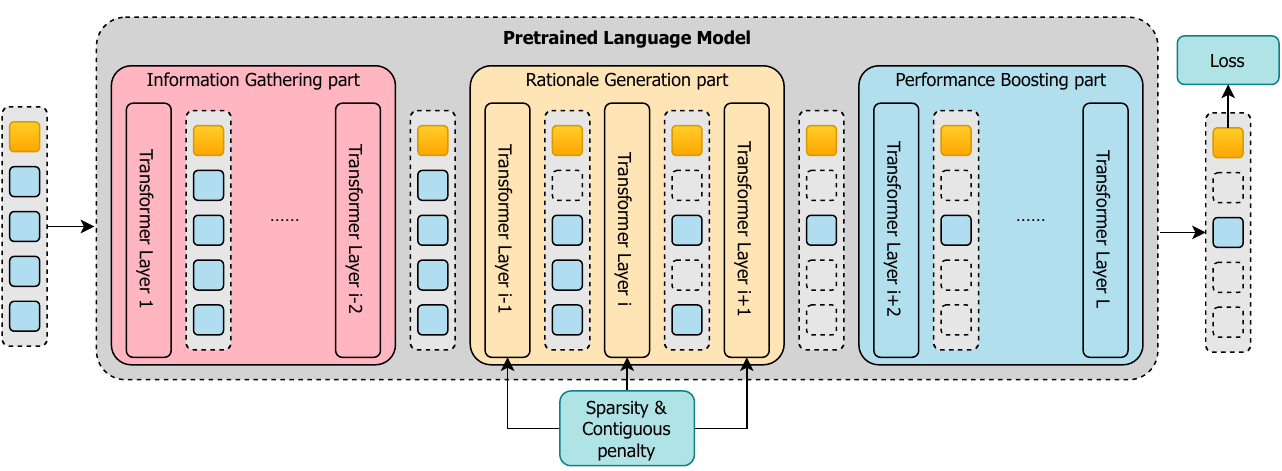}
    \caption{
    The framework of our approach~(YOFO). We split the pre-trained language model into three parts. The \textit{Information Gathering} part aims to gather the information from all the tokens and learn the importance of tokens. The \textit{Rationale Generation} part (3 layers in this picture) selects the important tokens to satisfy the aimed sparsity level. The \textit{Performance Boosting} part takes the important tokens from the previous part and enhances the performance of downstream tasks. Any pre-trained language model and token selection algorithm can be applied to this framework.
    }
    \label{fig:model}
\end{figure*}

Formally, we are given text-label pairs $\{\mathbf{x}, y\}$ from the dataset $\mathcal{D}$ and a pre-trained language model $\mathrm{PLM}$ with $N$ layers. The token embedding for each text can be obtained using Equation (\ref{eq:word embedding}).
\begin{gather}
    \label{eq:word embedding} \mathbf{W}=\mathrm{word\_embedding}(\mathbf{x}) \in \mathrm{R}^{L\times D}
\end{gather}
Here, $L$ and $D$ represent the input length and dimensions of the token embedding, respectively. We consider the token embedding $\mathbf{W}$ as the initial hidden states of the pre-trained language model $\mathbf{H}_0$. These hidden states will be fed into the following $N$ transformer layers.

To remove unimportant tokens in the $i$-th layer, we calculate the binary mask $\mathbf{m}_i \in \{0, 1\}^{L}$ for the $i$-th layer based on the corresponding hidden states $\mathbf{H}_{i-1}$.
This binary mask is then applied to the hidden states, resulting in meaningful representations $\mathbf{\tilde{H}}_{i-1}$ for the subsequent layer. 
Additionally, to ensure a monotonically decreasing token length as we go deeper into the layers, the hidden states also multiply the previously obtained masks. 
The process can be formulated as follows:
\begin{gather}
    \label{eq:mask} \mathbf{\tilde{m}}_i=g_i(\mathbf{H}_{i-1}) \\
    \label{eq:prod} \mathbf{m}_i=\mathbf{\tilde{m}}_i\odot\mathbf{m}_{i-1} \\
    \label{eq:apply} \mathbf{\tilde{H}}_{i-1}=\mathbf{H}_{i-1}\odot\mathbf{{m}}_i \\
    \label{eq:nlayer} \mathbf{H}_{i}=\mathrm{PLM}_i(\mathbf{\tilde{H}}_{i-1}) \text{, for }i\text{ in }1,2,...,N. 
\end{gather}
The function $g_i$ maps the hidden states in the $i$-th layer to a binary mask that takes values of either 0 or 1. In our approach, we employ a two-layer MLP and leverage the Gumbel softmax trick~\cite{gumbel} for this purpose.
$\mathbf{m}_{0}$ is set as a vector filled with ones.
Importantly, it should be noted that the classification token, such as [CLS] in BERT~\cite{JacobDevlin2018BERTPO}, which is used for later classification, is never deleted.

We can perform a sparsity constraint in the final layer of PLM to choose text snippets at a predefined sparsity level $s$. 
The loss term below will be added to the final loss:
\begin{gather}
    \label{eq:s-final} L_{s-final}=|\frac1L\sum_{j=1}^L m_N^j-s|
\end{gather}

\subsection{Contiguous Penalty}
We have noticed that when using sparse control only, the model tends to generate fragmented and unintelligible snippets. 
To address this issue, we introduce a contiguous penalty to encourage the generation of more coherent and readable snippets.
Formally, we aim to minimize the rate of change in the sequence length direction by adding a regularization term, denoted as $|m^i-m^{i-1}|$.

The overall loss function is defined as follows:
\begin{gather}
    \label{eq:contiguous} L_{contiguous}=\frac{\sum_{i=1}^N \sum_{j=1}^{L-1} |m_i^{j+1}-m_i^j|}{N(L-1)} \\
    \label{eq:overall_loss} L_{overall}=L_{task}+\beta L_{s-final}+\gamma L_{contiguous} 
\end{gather}
where $\beta$ and $\gamma$ are the balancing factors for different loss terms.

\subsection{Length Configuration}
\label{section:LC}
Directly imposing a sparsity constraint in the last layer has limitations as we lose detailed control of the token deletion. 
Our experiments in Section~\ref{sec:main_results} have shown that constraints in the last layer perform poorer than carefully designed token decay strategy.
In this section, we introduce a \textit{Length Configuration} $\{l_1, l_2, ..., l_N\} \in [0, 1]$ that allows for flexible control over the remaining tokens in each layer. 
Here, $l_i$ represents the percentage of tokens to be retained in the $i$-th layer. 
Importantly, the length configuration list is monotonically decreasing. 
We achieve this by adjusting the mask $\boldsymbol{m}_i$ to achieve the desired sparsity level $l_i$.
Based on the above assumption, we replace the sparsity penalty in Equation~(\ref{eq:s-final}) with a layer-wise sparsity control term as follows.
\begin{gather}
    \label{eq:sparsity} L_{sparsity}=\frac{1}{N}\sum_{i=1}^N |\frac1L\sum_{j=1}^L m_i^j-l_i|
\end{gather}

To gain a better understanding of how deleting tokens can affect model performance, we divide the pre-trained language models into three distinct parts as shown in Figure~\ref{fig:model}: \textit{Information Gathering} (IG), \textit{Rationale Generation} (RG), and \textit{Performance Boosting} (PB). 
In the IG part, all texts are retained ($l_i=1$) to gather token information.
In the RG part, the sparsity level is reduced to a predetermined value, while in the PB part, the sparsity level remains unchanged. 
To achieve this, we introduce three hyperparameters: $k\in [0, N]$ determines the number of layers where the text is fully retained for information gathering; $d\in [0, N-k]$ indicates the range from the $(k+1)$-th to the $(k+d)$-th layer where rationale should be generated; and $s\in [0,1]$ represents the sparsity level in the PB part. 
We have experimented with various methods to reduce the sparsity level in the RG part, including \textit{Cliff}, \textit{Linear}, \textit{Exponential}, and \textit{Logarithmic} decay. 
We use \textit{Cliff} decay as the default setting. 
A comparison of these token decay strategies is provided in Section \ref{sec:decay_compare}.

\subsection{Training}
In our preliminary experiments, we observed that directly multiplying the mask $\mathbf{m}_i$ with the hidden states $\mathbf{H}_i$ led to extremely poor performance during the training stage. 
We believe this degradation is due to the zeroing-out operation heavily disrupting the gradient flow through the hidden states.
To mitigate the negative impact of this observation, instead of directly zeroing out the hidden states, we now multiply the mask by the attention scores assigned to the tokens that should be zeroed out.
In the training stage, Equation (\ref{eq:apply}) and (\ref{eq:nlayer}) are replaced with Equation (\ref{eq:apply_attention}) and (\ref{eq:nlayer_attention}).
\begin{gather}
    \label{eq:apply_attention} \mathbf{\tilde{A}}_i^h=\mathbf{A}_i^h\odot\mathbf{{m}}_i \text{, for }h\text{ in }1,2,...,H. \\
    \label{eq:nlayer_attention} \mathbf{H}_i=\mathrm{PLM}_i(\mathbf{H}_{i-1};\mathbf{\tilde{A}}_i) \text{, for }i\text{ in }1,2,...,N. 
\end{gather}
where $\mathbf{A}_i^h$ represents the attention score of the $i$-th layer and the $h$-th head, and $H$ indicates the total number of heads in the multi-head attention mechanism.


\begin{table*}[]
\centering
\begin{tabular}{l|lllll|lllll|lllll}
\toprule
\multirow{2}{*}{Methods} & \multicolumn{5}{c|}{Appearance}                                                                                           & \multicolumn{5}{c|}{Aroma}                                                                                                & \multicolumn{5}{c}{Palate}                                                                                                \\ \cline{2-16} 
                         & \multicolumn{1}{c}{S} & \multicolumn{1}{c}{ACC} & \multicolumn{1}{c}{P} & \multicolumn{1}{c}{R} & \multicolumn{1}{c|}{F1} & \multicolumn{1}{c}{S} & \multicolumn{1}{c}{ACC} & \multicolumn{1}{c}{P} & \multicolumn{1}{c}{R} & \multicolumn{1}{c|}{F1} & \multicolumn{1}{c}{S} & \multicolumn{1}{c}{ACC} & \multicolumn{1}{c}{P} & \multicolumn{1}{c}{R} & \multicolumn{1}{c}{F1} \\ \hline
RNP*                      & 18.7                  & 84.0                       & 72.0                    & 72.7                  & 72.3                    & 15.1                  & 85.2                       & 59.0                 & 57.2                  & 58.1                    & 13.4                    & 90.0                       & 63.1                  & 68.2                  & 65.5                   \\
DMR*                      & 18.2                  & -                       & 71.1                  & 70.2                  & 70.7                    & 15.4                  & -                       & 59.8                  & 58.9                  & 59.3                    & 11.9                  & -                       & 53.2                  & 50.9                  & 52.0                   \\
A2R*                      & 18.4                  & 83.9                       & 72.7                  & 72.3                  & 72.5                    & 15.4                  & 86.3                       & 63.6                  & 62.9                  & 63.2                    & 12.4                  & 81.2                       & 57.4		
                  & 57.3                  & 57.4                   \\
FR**                       & 18.4                  & \ul{87.2}                    & 82.9                  & 82.6                  & 82.8                    & 15.0                  & \textbf{88.6}                    & 74.7                  & 72.1                  & 73.4                    & 12.1                    & \textbf{89.7}                    & 67.8                  & 66.2                  & 67.0                   \\
MGR*                       & 18.4                  & 86.1                    & 83.9                  & 83.5                  & 83.7                      & 15.6                  & 86.6                    & 76.6                  & 76.5                  & 76.5                    & 12.4                  & 85.1                    & 66.6                  & 66.6                  & 66.6                   \\
DR*                       & 18.6                  & 85.3                    & 84.3                  & 84.8                  & 84.5                      & 15.6                  & \ul{87.2}                    & 77.2                  & 77.5                  & 77.3                    & 13.3                  & 85.7                    & 65.1                  & 69.8                  & 67.4                   \\  \hline \hline
BERT-RNP               &18.8&	86.4&	83.2&	82.3&	82.8&	15.2&	82.8&	77.4&	70.8&	73.9&	13.0&	87.3&	64.5&	63.1&	63.8
                   \\
BERT-FR        &21.3&	84.2&	24.2&	27.2&	25.6&	17.9&	83.8	&51.6	&55.8&	53.6&	12.2&	88.0&	57.2&	52.6	&54.8        
                   \\ \hline\hline
YOFO-final               & 18.4                  & \textbf{87.4}           & \ul{90.9}         & \textbf{88.0}         & \textbf{89.5}           & 14.7                  & 85.7           & \ul{92.9}         & \ul{82.5}         & \ul{87.4}           & 13.6                  & 87.8           & \ul{72.2}         & \ul{73.5}         & \ul{72.8}         \\ 
YOFO               & 18.1                  & 85.6           & \textbf{91.3}         & \ul{87.1}         & \ul{89.2}           & 15.4                  & 86.8           & \textbf{94.3}         & \textbf{87.9}         & \textbf{91.0}           & 13.2                  & \ul{88.4}           & \textbf{79.5}         & \textbf{79.0}         & \textbf{79.2}         \\ 
\bottomrule
\end{tabular} 
\caption{Results on the high-sparse decorrelated BeerAdvocate dataset~\cite{beer}. "*" represents the results obtained from~\cite{DR} directly. \textbf{Bold} indicates the best results among different methods, while \ul{underline} indicates the second-place results among different methods.}
\label{table:high_sparse_beer}
\end{table*}

\section{Experiments}
\subsection{Experimental Setup}
\paragraph{Datasets}
We performed experiments on two commonly used unsupervised rationale extraction datasets: BeerAdvocate\cite{beer} and the Hotel Review dataset~\cite{hotel}. 

The BeerAdvocate dataset~\cite{beer} is a multi-aspect sentiment prediction dataset. 
It consists of texts along with corresponding aspect scores ranging from 0 to 1, including aspects such as \textit{appearance}, \textit{aroma}, and \textit{palate}.
The training and development sets do not have labeled rationales, but the test set contains 994 samples with rationale annotations for all aspects.
Notably, the scores across different aspects within the same sample exhibit high correlation, resulting in highly spurious correlations.
For the BeerAdvocate dataset, we conducted experiments on the decorrelated version proposed by ~\cite{lei-etal-2016-rationalizing}.
We binarized the dataset into binary classification tasks using a positive threshold of 0.6 and a negative threshold of 0.4~\cite{bao2018deriving}.
We run our model, YOFO, on two sparsity levels: high-sparse and low-sparse. 
In the high-sparse decorrelated dataset, the sparsity level approximates the golden sparsity observed in the test set.
In the low-sparse decorrelated dataset, the sparsity level is comparatively lower but allows for convenient comparisons with previous works.
To examine the susceptibility of our model to spurious correlations, we also utilized the correlated BeerAdvocate dataset from ~\cite{MGR}.

The Hotel Review dataset~\cite{hotel} is another widely used dataset for multi-aspect sentiment classification and rationale extraction. It includes texts along with three aspect labels: \textit{location}, \textit{service}, and \textit{cleanliness}. 
In addition to the aspect labels, the test set of this dataset also provides rationale annotations for all three aspects, with 200 samples. Since the original labels are on a scale of 0 to 5 stars, we utilize the binarized version proposed by ~\cite{bao2018deriving}.
For the Hotel Review dataset, we only conducted a low-sparse experiment as the golden sparsity level is relatively low, at around 10\%.

\paragraph{Baselines}
We compared the performance of our models with several state-of-the-art baselines on the BeerAdvocate and Hotel Review datasets. 
These baselines, including RNP~\cite{lei-etal-2016-rationalizing}, HardKuma~\cite{bastings-etal-2019-interpretable}, CAR~\cite{CAR}, INVRAT~\cite{INVRAT}, DMR~\cite{DMR}, FR~\cite{FR}, Inter-RAT~\cite{inter-rat}, MGR~\cite{MGR}, and DR~\cite{DR}, were discussed in Section \ref{section:related_rationale}.
Our methods, YOFO with the default setting (YOFO) and YOFO that only constrains sparsity in the final layer (YOFO-final), were used for comparison. 
In order to ensure a fair comparison, we implemented the following baselines since we utilized a pre-trained language model in YOFO.
BERT-RNP is the RNP framework equipped with BERT, which serves as our direct baseline. 
BERT-FR is similar to BERT-RNP, except that the parameters of the generator and predictor are shared.

\paragraph{Metrics}
We will use token-level f1 to measure the quality of extracted rationales and accuracy for downstream tasks, following previous works~\cite{FR, DMR}.
In our result tables, we define S as the sparsity level of selected rationales, computed using the formula $S=\frac{\#selected\text{ }tokens}{\#tokens}$. 
P, R, and F1 represent precision, recall, and F1 score for rationale extraction, respectively. 
ACC and Val ACC denote the accuracy of the test and validation sets, respectively, for the downstream tasks. 
For more information about our implementation details, please refer to Appendix~\ref{sec:imp_main}.

\begin{table*}[]
\centering
\begin{tabular}{l|lllll|lllll|lllll}
\toprule
\multirow{2}{*}{Methods} & \multicolumn{5}{c|}{Appearance}                                                                                           & \multicolumn{5}{c|}{Aroma}                                                                                                & \multicolumn{5}{c}{Palate}                                                                                                \\ \cline{2-16} 
                         & \multicolumn{1}{c}{S} & \multicolumn{1}{c}{ACC} & \multicolumn{1}{c}{P} & \multicolumn{1}{c}{R} & \multicolumn{1}{c|}{F1} & \multicolumn{1}{c}{S} & \multicolumn{1}{c}{ACC} & \multicolumn{1}{c}{P} & \multicolumn{1}{c}{R} & \multicolumn{1}{c|}{F1} & \multicolumn{1}{c}{S} & \multicolumn{1}{c}{ACC} & \multicolumn{1}{c}{P} & \multicolumn{1}{c}{R} & \multicolumn{1}{c}{F1} \\ \hline
RNP*                      & 11.9                  & -                       & 72.0                    & 46.1                  & 56.2                    & 10.7                  & -                       & 70.5                  & 48.3                  & 57.3                    & 10.0                    & -                       & 53.1                  & 42.8                  & 47.5                   \\
CAR*                      & 11.9                  & -                       & 76.2                  & 49.3                  & 59.9                    & 10.3                  & -                       & 50.3                  & 33.3                  & 40.1                    & 10.2                  & -                       & 56.6                  & 46.2                  & 50.9                   \\
DMR*                      & 11.7                  & -                       & 83.6                  & 52.8                  & 64.7                    & 11.7                  & -                       & 63.1                  & 47.6                  & 54.3                    & 10.7                  & -                       & 55.8                  & 48.1                  & 51.7                   \\
FR**                       & 12.7                  & 83.9                    & 77.6                  & 53.3                  & 63.2                    & 10.8                  & \textbf{87.6}                    & 82.9                  & 57.9                  & 68.2                    & 10.0                    & 84.5                    & 69.3                  & 55.8                  & 61.8                   \\
DR*                       & 11.9                  & 81.4                    & 86.8                  & 55.9                  & 68.0                      & 11.2                  & 80.5                    & 70.8                  & 57.1                  & 63.2                    & 10.5                  & 81.4                    & 71.2                  & 60.2                  & 65.3                   \\ \hline\hline
BERT-RNP                 & 13.8                     & 84.9                    & 84.7                  & 61.4                  & 71.2                    & 13.9                  & 86.2                    & 70.9                 & 59.5                  & 64.7                    & 11.9                  & 85.1                    & 73.1                  & 65.5                  & 69.1                    \\
BERT-FR                  & 15.4 & 	84.4 &	22.8 &	18.5& 	20.4&	14.4&	85.5&	21.4&	18.7&	19.9&	14.1&	86.6&	16.6&	17.5&	17.1
                   \\ \hline\hline
YOFO-final               & 13.2                  & \textbf{87.5}           & \textbf{97.4}         & \textbf{67.6}         & \textbf{79.8}           & 12.4                  & 85.5           & \textbf{95.4}         & \textbf{71.2}         & \textbf{81.6}           & 10.2                  & \ul{87.6}           & \ul{83.3}         & \ul{63.7}         & \ul{72.2}         \\
YOFO               & 13.1                  & \ul{87.0}           & \ul{97.1}         & \ul{66.9}         & \ul{79.2}           & 12.1                  & \ul{86.3}           & \ul{94.1}         & \ul{68.9}         & \ul{79.5}           & 10.9                  & \textbf{87.8}           & \textbf{88.5}         & \textbf{72.7}         & \textbf{79.8}         \\
\bottomrule
\end{tabular}
\caption{The results of different methods on low-sparse decorrelated BeerAdvocate dataset~\cite{beer}. "*" and "**" represent the results obtained from~\cite{DR}and~\cite{FR}, respectively. \textbf{Bold} indicates the best results among different methods.  \ul{underline} indicates the second-place results among different methods.}
\label{table:low_sparse_beer}
\end{table*}


\begin{table*}[]
\centering
\begin{tabular}{l|lllll|lllll|lllll}
\toprule
\multicolumn{1}{c|}{\multirow{2}{*}{Methods}} & \multicolumn{5}{c|}{Location}                                                                                                    & \multicolumn{5}{c|}{Service}                                                                                                     & \multicolumn{5}{c}{Cleanliness}                                                                                                 \\ \cline{2-16} 
\multicolumn{1}{c|}{}                         & \multicolumn{1}{c}{S} & \multicolumn{1}{c}{ACC} & \multicolumn{1}{c}{P} & \multicolumn{1}{c}{R} & \multicolumn{1}{c|}{F1} & \multicolumn{1}{c}{S} & \multicolumn{1}{c}{ACC} & \multicolumn{1}{c}{P} & \multicolumn{1}{c}{R} & \multicolumn{1}{c|}{F1} & \multicolumn{1}{c}{S} & \multicolumn{1}{c}{ACC} & \multicolumn{1}{c}{P} & \multicolumn{1}{c}{R} & \multicolumn{1}{c}{F1} \\ \hline
RNP**                                           & 8.8                          & 97.5                    & 46.2                  & 48.2                  & 47.1                    & 11.0                           & \ul{97.5}                    & 34.2                  & 32.9                  & 33.5                    & 10.5                         & 96.0                      & 29.1                  & 34.6                  & 31.6                   \\
DMR*                                           & 10.7                         & -                       & 47.5                  & 60.1                  & 53.1                    & 11.6                         & -                       & 43.0                    & 43.6                  & 43.3                    & 10.3                         & -                       & 31.4                  & 36.4                  & 33.7                   \\
A2R*                                           & 8.5                          & 87.5                    & 43.1                  & 43.2                  & 43.1                    & 11.4                         & 96.5                    & 37.3                  & 37.2                  & 37.2                    & 8.9                          & 94.5                    & 33.2                  & 33.3                  & 33.3                   \\
FR*                                            & 9.0                            & 93.5                    & 55.5         & 58.9                  & 57.1           & 11.5                         & 94.5                    & 44.8                  & 44.7                  & 44.8                    & 11.0                           & 96.0                      & 34.9                  & 43.4                  & 38.7                   \\
MGR**                                           & 9.7                          & 97.5                    & 52.5                  & 60.5                  & 56.2                    & 11.8                         & 96.5                    & 45.0                    & 46.4                  & 45.7                    & 10.5                         & 96.5                    & 37.6                  & 44.5                  & 40.7                   \\
DR*                                            & 9.6                          & 96.5                    & 53.6                  & \ul{60.9}         & 57.0                      & 11.5                         & 96.0                      & 47.1                  & 47.4                  & 47.2                    & 10.0                           & \ul{97.0}                      & 39.3                  & 44.3                  & 41.8                   \\ \hline\hline
BERT-RNP & 9.4 & \ul{98.0} & 47.2 & 49.9 & 48.5 & 10.2 & \textbf{99.5} & 39.1 & 33.1  & 35.8 & 9.8 & \textbf{100.0} & 45.1 & 45.2 & 45.2 \\
BERT-FR  & 8.5 & \textbf{98.5}   & 14.3   & 13.9 & 14.1 & 10.8 & \textbf{99.5} & 26.4 & 23.8 & 25.0 & 11.1 & \textbf{100.0} & 25.6 & 29.4 & 27.4 \\ \hline\hline
YOFO-final                                          & 10.9                          & \ul{98.0}           & \ul{54.2}                  & \textbf{64.1}                  & \textbf{58.8}                    & 10.5                         & \textbf{99.5}           & \textbf{60.8}         & \textbf{55.9}         & \textbf{58.3}             & 9.3 & \textbf{100.0} & 47.5 & 45.2 & 46.3 \\
YOFO                                          & 9.7                          & \ul{98.0}           & \textbf{55.7}                  & 60.4                  & \ul{58.0}                    & 11.9                         & \textbf{99.5}           & \ul{58.3}         & \ul{57.4}         & \ul{57.9}             & 10.6                         & \textbf{100.0}            & \textbf{49.9}         & \textbf{54.4}         & \textbf{52.1} \\ 
\bottomrule
\end{tabular}
\caption{The results of different methods on Hotel Review dataset~\cite{hotel}. "*" and "**" represent the results obtained from~\cite{DR}and~\cite{MGR}, respectively. \textbf{Bold} indicates the best results among different methods.  \ul{underline} indicates the second-place results among different methods. }
\label{table:hotel}
\end{table*}

\subsection{Main Results}
\label{sec:main_results}

\subsubsection{High-sparse decorrelated BeerAdvocate dataset}
The results on the high-sparse, decorrelated BeerAdvocate dataset are shown in Table~\ref{table:high_sparse_beer}. The table reveals the following corollary:

1) YOFO-series methods outperform previous state-of-the-art and PLM baselines (BERT-RNP and BERT-FR), securing first and second places. Specifically, YOFO achieves a performance gain of 4.7\%, 13.7\%, and 11.8\% higher in token-level F1 for three aspects, respectively, and 5.0\%, 10.1\%, and 5.4\% for YOFO-final. The superior performances demonstrate the effectiveness of our method.

2) YOFO-final performs slightly better than YOFO in the Appearance aspect, but worse in other aspects, particularly with a 6.4\% drop in the Palate aspect. We attribute this to the relative difficulty of the final aspect and the lack of sparsity controls within the YOFO-final model, which hinders the faithful identification of important tokens.

3) YOFO-series methods exhibit slightly lower ACC scores compared to SOTA methods. This is due to overfitting caused by training the model for 10 epochs on the dataset. We will demonstrate that in Section~\ref{sec:case_layer}.

4) BERT-RNP outperforms BERT-FR by a significant margin, which is opposite to the discovery in~\cite{FR}. This suggests the ineffectiveness of the FR (FR, Reference) mechanism, which shares the encoder's parameters between the generator and predictor, at least when using PLMs in the RNP framework.

\begin{table*}[]
\centering
\begin{tabular}{l|l|llll|llll|llll}
\toprule
\multicolumn{1}{c|}{\multirow{2}{*}{Methods}} & \multicolumn{1}{c|}{\multirow{2}{*}{S}} & \multicolumn{4}{c|}{Appearance}                                                                   & \multicolumn{4}{c|}{Aroma}                                                                        & \multicolumn{4}{c}{Palate}                                                                       \\ \cline{3-14} 
\multicolumn{1}{c|}{}                         & \multicolumn{1}{c|}{}                   & \multicolumn{1}{c}{ACC} & \multicolumn{1}{c}{P} & \multicolumn{1}{c}{R} & \multicolumn{1}{c|}{F1} & \multicolumn{1}{c}{ACC} & \multicolumn{1}{c}{P} & \multicolumn{1}{c}{R} & \multicolumn{1}{c|}{F1} & \multicolumn{1}{c}{ACC} & \multicolumn{1}{c}{P} & \multicolumn{1}{c}{R} & \multicolumn{1}{c}{F1} \\ \hline
RNP*                                           & \multirow{6}{*}{10}                     & -                       & 32.4                  & 18.6                  & 23.6                    & -                       & 44.8                  & 32.4                  & 37.6                    & -                       & 24.6                  & 23.5                  & 24.0                     \\
HardKuma*                                      &                                         & -                       & 53.6                  & 28.7                  & 37.4                    & -                       & 29.3                  & 25.9                  & 27.3                    & -                       & 7.7                   & 6.0                     & 6.8                    \\
INVRAT*                                        &                                         & -                       & 42.6                  & 31.5                  & 36.2                    & -                       & 41.2                  & 39.1                  & 40.1                    & -                       & 34.9                  & 45.6                  & 39.5                   \\
Inter-RAT*                                     &                                         & -                       & 66.0                    & 46.5                  & 54.6                    & -                       & 55.4                  & 47.5                  & 51.1                    & -                       & 34.6                  & 48.2                  & 40.2                   \\
MGR**                                           &                                         & 80.5                    & 87.5                  & 51.7                  & 65.0                      & 89.7                       & 78.7                  & 52.2                  & 62.8                    & 86.0                      & 65.6         & 57.1                  & 61.1                   \\
YOFO(\textit{ours})                                    &                                         & \textbf{87.7}           & \textbf{96.4}         & \textbf{61.9}         & \textbf{75.4}           & \textbf{92.7}           & \textbf{95.4}         & \textbf{65.2}         & \textbf{77.5}           & \textbf{91.9}           & \textbf{67.4}                  & \textbf{67.4}         & \textbf{67.4}          \\ \hline
RNP*                                           & \multirow{6}{*}{20}                     & -                       & 39.4                  & 44.9                  & 42.0                      & -                       & 37.5                  & 51.9                  & 43.5                    & -                       & 21.6                  & 38.9                  & 27.8                   \\
HardKuma*                                      &                                         & -                       & 64.9                  & 69.2                  & 67.0                      & -                       & 37.0                    & 55.8                  & 44.5                    & -                       & 14.6                  & 22.3                  & 17.7                   \\
INVRAT*                                        &                                         & -                       & 58.9                  & 67.2                  & 62.8                    & -                       & 29.3                  & 52.1                  & 37.5                    & -                       & 24.0                    & 55.2                  & 33.5                   \\
Inter-RAT*                                     &                                         & -                       & 62.0                    & 76.7                  & 68.6                    & -                       & 44.2                  & 65.4                  & 52.8                    & -                       & 26.3                  & 59.1                  & 36.4                   \\
MGR**                                           &                                         & 85.6                    & 76.3         & 83.6                  & 79.8                    & 89.6                    & 64.4                  & 81.3                  & 71.9                    & 89.3                    & \textbf{47.1}         & 73.1                  & \textbf{57.3}          \\
YOFO(\textit{ours})                                    &                                         & \textbf{88.4}           & \textbf{77.5}                  & \textbf{87.6}         & \textbf{82.2}             & \textbf{91.9}           & \textbf{78.7}         & \textbf{92.8}         & \textbf{85.2}           & \textbf{91.3}           & 44.6                    & \textbf{75.4}         & 56.0                   \\ \hline
RNP*                                           & \multirow{6}{*}{30}                     & -                       & 24.2                  & 41.2                  & 30.5                    & -                       & 27.1                  & 55.7                  & 36.4                    & -                       & 15.4                  & 42.2                  & 22.6                   \\
HardKuma*                                      &                                         & -                       & 42.1                  & 82.4                  & 55.7                    & -                       & 24.6                  & 57.7                  & 34.5                    & -                       & 21.7                  & 49.7                  & 30.2                   \\
INVRAT*                                        &                                         & -                       & 41.5                  & 74.8                  & 53.4                    & -                       & 22.8                  & 65.1                  & 33.8                    & -                       & 20.9                  & 71.6                  & 32.3                   \\
Inter-RAT*                                     &                                         & -                       & 48.1                  & 82.7                  & 60.8                    & -                       & 37.9                  & 72.0                    & 49.6                    & -                       & 21.8                  & 66.1                  & 32.8                   \\
MGR**                                           &                                         & 88.5                    & 57.2                  & 93.9         & 71.1                    & 91.6                    & 45.8                  & 87.4         & 60.1                    & 89.3                    & 27.3                  & 66.5                  & 38.7                   \\
YOFO(\textit{ours})                                    &                                         & \textbf{88.9}           & \textbf{63.5}         & \textbf{94.3}                  & \textbf{75.9}           & \textbf{92.4}           & \textbf{53.6}         & \textbf{88.7}                  & \textbf{66.8}           & \textbf{91.6}           & \textbf{34.0}         & \textbf{75.7}         & \textbf{46.9}         \\
\bottomrule
\end{tabular}
\caption{The results of different methods on correlated BeerAdvocate Dataset~\cite{beer}. "*" and "**" represent the results obtained from ~\cite{inter-rat} and ~\cite{MGR}, respectively. \textbf{Bold} indicates the best results among all methods in the same setting. }
\label{table:correlate}
\end{table*}

\subsubsection{Low-sparse datasets}
The results on the low-sparse, decorrelated BeerAdvocate dataset are presented in Table~\ref{table:low_sparse_beer}. Similar conclusions can be drawn, but with a few modifications:

1) YOFO-final surpasses YOFO in the first two aspects.

2) YOFO-series methods demonstrate greater advantages in the low-sparse scenario. YOFO achieves 11.2\%, 11.3\%, and 14.5\% higher token-level F1 scores than previous state-of-the-art (SOTA) methods in three aspects, respectively, and 11.2\%, 18.4\%, and 6.9\% for YOFO-final. 

3) Interestingly, our implemented BERT-RNP method outperforms GRU-based methods in the Appearance and Palate aspects. This indicates the potential of deploying BERT-like pre-trained language models within the RNP framework.


Finally, the results on the Hotel Review dataset are presented in Table~\ref{table:hotel}. Similar conclusions can be drawn, except for higher accuracy compared to state-of-the-art approaches. 
The model performs well on downstream tasks but performs poorly on rationale extraction. A more in-depth analysis is shown in Section~\ref{sec:case_hotel}.

\subsection{Results on Correlated BeerAdvocate dataset}
To demonstrate that our model has less impact from spurious correlations, we run our model on the correlated BeerAdvocate dataset~\cite{beer}. The results are presented in Table~\ref{table:correlate}. YOFO outperforms all baselines, often by a significant margin. However, when the sparsity level is 20\%, we perform slightly worse than the best model MGR~\cite{MGR} in the Palate aspect.
YOFO demonstrates a substantial lead in test accuracy compared to all baselines, particularly at low sparsity levels. This suggests the power of PLMs and the effectiveness of YOFO.

The results provide evidence that our model effectively mitigates the issue of spurious correlations. The reason for YOFO's superior performance over other RNP-based methods is as follows:
RNP-based methods directly predicted labels using generated rationales, often selecting "useless but highly correlated" text. In contrast, YOFO takes a different approach. Its predictions do not depend directly on the rationales. Instead, rationales support YOFO's predictions.

\subsection{Results on Skewed Decorrelated BeerAdvocate dataset}
To further prove that our model never suffers from interlocking. 
We conducted experiments on the skewed decorrelated BeerAdvocate dataset. 
Further details on the experiments can be found in Appendix~\ref{sec:imp_skew}.

The results, as shown in Table~\ref{table:skewd}, reveal that RNP suffers significantly from interlocking problems in the Aroma aspect, particularly in the skew20 scenario, while our model experiences minimal interlocking issues and achieves the best results. In the more challenging Palate aspect, RNP nearly fails, while DR's performance declines as the skew epoch increases. However, YOFO maintains its performance and does not suffer significant performance degradation, further demonstrating that YOFO does not suffer from the interlocking problem.
\begin{table}[htbp!]
\begin{tabular}{l|l|llll|llll|llll}
\toprule
\multirow{2}{*}{Aspect} & \multirow{2}{*}{setting} & \multicolumn{4}{c|}{RNP~\cite{lei-etal-2016-rationalizing}}  & \multicolumn{4}{c|}{DR~\cite{DR}}   & \multicolumn{4}{c}{YOFO}                                      \\ \cline{3-14} 
                        &                          & Acc  & P    & R    & F1   & Acc  & P    & R    & F1   & Acc           & P             & R             & F1            \\ \hline
\multirow{3}{*}{Aroma}  & skew10                   & 82.6 & 68.5 & 63.7 & 61.5 & 85.0 & 77.3 & 75.7 & 76.5 & \textbf{87.5} & \textbf{96.6} & \textbf{82.5} & \textbf{89.0} \\
                        & skew15                   & 80.4 & 54.5 & 51.6 & 49.3 & 85.4 & 76.1 & 77.2 & 76.6 & \textbf{85.6} & \textbf{95.8} & \textbf{84.2} & \textbf{89.6} \\
                        & skew20                   & 76.8 & 10.8 & 14.1 & 11.0 & 85.5 & 77.3 & 76.2 & 76.8 & \textbf{86.4} & \textbf{94.9} & \textbf{85.1} & \textbf{89.7} \\ \hline
\multirow{3}{*}{Palate} & skew10                   & 77.3 & 5.6  & 7.4  & 5.5  & 85.8 & 67.7 & 68.6 & 68.2 & \textbf{87.0} & \textbf{81.2} & \textbf{74.2} & \textbf{77.5} \\
                        & skew15                   & 77.1 & 1.2  & 2.5  & 1.3  & 83.9 & 66.3 & 66.7 & 66.5 & \textbf{87.3} & \textbf{82.6} & \textbf{77.5} & \textbf{79.9} \\
                        & skew20                   & 75.6 & 0.4  & 1.4  & 0.6  & 85.0 & 59.4 & 62.6 & 61.0 & \textbf{86.7} & \textbf{81.2} & \textbf{76.2} & \textbf{78.6} \\
\bottomrule
\end{tabular}
\caption{Results on skewed decorrelated BeerAdvocate Dataset~\cite{beer} }
\label{table:skewd}
\end{table}

\section{Analysis}
YOFO demonstrates excellent performance across all the aforementioned experimental settings. However, we need to delve deeper into understanding the nature of its rationales and how to precisely control the sparsity level.
In this section, we will conduct a comprehensive analysis of YOFO and its rationales.
We will explore the significance of generated rationales and examine their characteristics in Section~\ref{sec:important}, \ref{sec:case_beer}, and \ref{sec:case_hotel}. 
Moreover, we will delve into a detailed layer-wise analysis and the token decay strategies in Section~\ref{sec:case_layer} and ~\ref{sec:decay_compare}.
\subsection{Are Generated Rationales by YOFO Important ?}
\label{sec:important}
We extracted rationales using YOFO on the train and dev sets of the BeerAdvocate Appearance aspect, referred to as the "rationale train set" and "rationale dev set," respectively. We trained a BERT model using the "rationale train set" and evaluated its performance on the "rationale dev set." Additionally, we established a "majority" baseline that always predicts the majority class in the dev set. The results are presented in Table~\ref{table:important}.
It is evident from the table that the model trained with generated rationales tremendously outperforms the simple "majority" baseline and has a 3.3\% decrease in accuracy compared to models trained and evaluated on full texts.
This indicates that the rationales generated by YOFO are important but not perfect.

\begin{table}[htbp!]
\centering
\begin{tabular}{l|l|l|l}
\toprule
& full texts & generated rationales & majority \\ \hline
Val ACC & 90.1 & 86.8 & 75.9 \\
\bottomrule
\end{tabular}
\caption{The performance on BeerAdvocate Appearance aspect.}
\label{table:important}
\end{table}





\begin{table*}[]
\centering
\begin{tabular}{p{3.5cm}|p{3.5cm}|p{3.5cm}|p{3.5cm}} 
\toprule
\hline
RNP~\cite{lei-etal-2016-rationalizing}                                                                                                                                                                                                                                                                                                                                                      & FR~\cite{FR}                                                                                                                                                                                                                                                                                                                                                                            & MGR~\cite{MGR}                                                                                                                                                                                                                                                                                                                                                                                  & YOFO~(\textit{ours})                                                                                                                                                                                                                                                                                                                                                                                          \\ 
\cline{1-4}
\textbf{Aspect}: Appearance                                                                                                                                                                                                                                                                                                                              & \textbf{\textbf{Aspect}}: Appearance                                                                                                                                                                                                                                                                                                                                          & \textbf{\textbf{Aspect}}: Appearance                                                                                                                                                                                                                                                                                                                                                 & \textbf{\textbf{Aspect}}: Appearance                                                                                                                                                                                                                                                                                                                                                           \\
\textbf{Label}:\textbf{~}Positive                                                                                                                                                                                                                                                                                               & \textbf{\textbf{Label}}:\textbf{\textbf{~}}Positive                                                                                                                                                                                                                                                                                       & \textbf{\textbf{Label}}:\textbf{\textbf{~}}Positive                                                                                                                                                                                                                                                                                       & \textbf{\textbf{Label}}:\textbf{\textbf{~}}Positive    \\   
\textbf{Pred}:\textbf{~}Negative                                                                                                                                                                                                                                                                                               & \textbf{\textbf{Pred}}:\textbf{\textbf{~}}Negative                                                                                                                                                                                                                                                                                       & \textbf{\textbf{Pred}}:\textbf{\textbf{~}}Negative                                                                                                                                                                                                                                                                                       & \textbf{\textbf{Pred}}:\textbf{\textbf{~}}Positive
\\
\textbf{Text}:~\textcolor{blue}{10 / 3 / 2009 500ml bottled aug 31 / 09 . \ul{pours}}\ul{ a hazy orange , and this thing is damn delicious .} incredibly easy to drink and so damn good . not a hint of the alc . \% at all . goes down so smooth . i loved the hints ofgrapefruit and bitterness to this beauty . worth the hype , and then some . & \textbf{\textbf{Text}}:~\textcolor{red}{10 / 3 / 2009 500ml bottled aug 31 / 09 . \ul{pours}}\ul{ a hazy orange , and this thing is damn delicious .} incredibly easy to drink and so damn good . not a hint of the alc . \% at all . goes down so smooth . i loved the hints of grapefruit and bitterness to this beauty . worth the hype , and then some . & \textbf{\textbf{Text}}:~\textcolor{pink}{10 / 3 / 2009 500ml bottled aug} 31 / 09 . \ul{pours}\ul{ a hazy orange , and this thing is damn delicious .} incredibly easy to drink and so damn good . not a hint of the alc . \% at all . goes down so smooth . i loved the hints of grapefruit and bitterness to this beauty . worth the hype , and then some . & \textbf{Text}:~{10 / 3 / 2009 500ml bottled aug} 31 / 09 . \textcolor{cyan}{\ul{pours a hazy orange , and this thing is damn delicious .}} incredibly easy to drink and so damn good . not a hint of the alc . \% at all . goes down so smooth . i loved the hints of grapefruit and bitterness to this beauty . worth the hype , and then some . \\
\hline
\bottomrule
\end{tabular}
\caption{Case studies on Appearance aspect of Beer Review dataset~\cite{beer}. We compare rationales generated by 4 different methods. Golden rationales are marked using \ul{underline}. Rationales from RNP~\cite{lei-etal-2016-rationalizing}, FR~\cite{FR}, MGR~\cite{MGR} and YOFO~(\textit{ours}) are highlighted using \textcolor{blue}{blue}, \textcolor{red}{red}, \textcolor{pink}{pink} and \textcolor{cyan}{cyan}, respectively.}
\label{table:decode_0}
\end{table*}

\subsection{Case Studies on BeerAdvocate Dataset}
\label{sec:case_beer}
\subsubsection{Right cases}
As shown in Table~\ref{table:decode_0}.
YOFO accurately predicted the correct label and selected the entire golden rationale, while the other three methods made incorrect predictions and chose rationales with limited overlap with the golden one.
This issue can be attributed to the interlocking problem, where the generator selects meaningless rationales and the predictor subsequently overfits to them.
In YOFO, the generator and predictor are the same model and do not have a sequential dependency like RNP.
Therefore, YOFO avoids the interlocking problem that is very serious in the RNP framework.

\begin{figure}
	\centering
	\subfloat[RNP]{\includegraphics[width=.43\columnwidth]{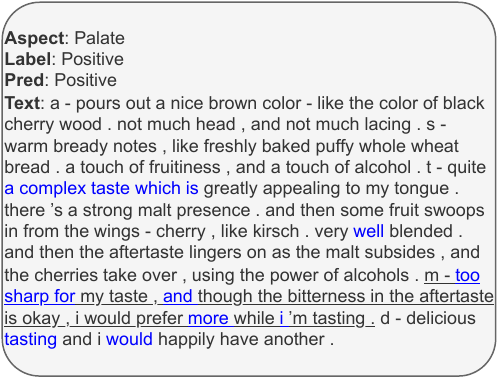}}\label{fig:RNP_case}
	\subfloat[FR]{\includegraphics[width=.43\columnwidth]{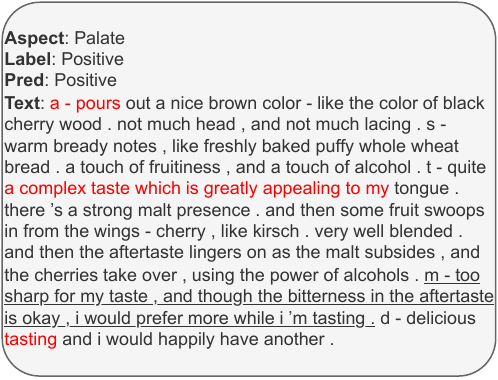}}\label{fig:FR_case} \\ 
	\subfloat[MGR]{\includegraphics[width=.43\columnwidth]{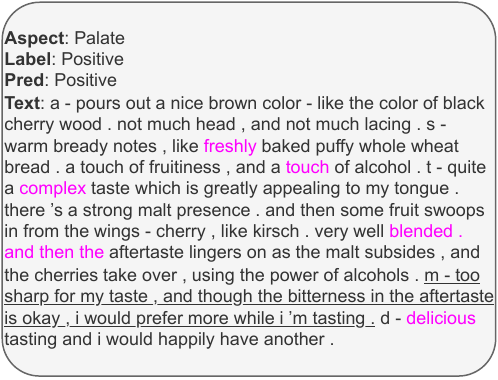}}\label{fig:MGR_case}
	\subfloat[YOFO(\textit{ours})]{\includegraphics[width=.43\columnwidth]{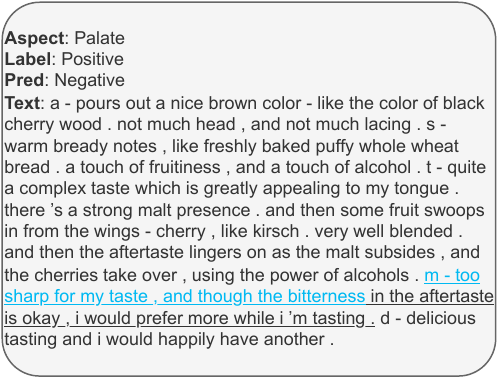}}\label{fig:YOFO_case}
	\caption{Case studies on Palate aspect of Beer Review dataset~\cite{beer}. We compare rationales generated by 4 different methods. Golden rationales are marked using \ul{underline}. Rationales from RNP~\cite{lei-etal-2016-rationalizing}, FR~\cite{FR}, MGR~\cite{MGR} and YOFO~(\textit{ours}) are highlighted using \textcolor{blue}{blue}, \textcolor{red}{red}, \textcolor{pink}{pink} and \textcolor{cyan}{cyan}, respectively.}
\label{fig:decode_1}
\end{figure}

\subsubsection{Error cases}
In Figure~\ref{fig:decode_1}, YOFO makes an incorrect label prediction, whereas the other methods make correct predictions. However, the generated rationales from these three methods perform poorly and still fall into the interlocking problem. 
MGR's rationale includes the term "delicious", which is not present in the golden rationale, yet it still leads to accurate predictions. Similarly, the remaining two methods heavily rely on the phrase "complex taste which is greatly appealing to my tongue", which is positive but not part of the golden rationale. 
On the other hand, YOFO includes a portion of the golden rationale but makes an incorrect prediction. 
However, "Too sharp for my taste" is actually a negative statement. 
The reason behind YOFO's erroneous prediction lies in its failure to capture the latter part of the sentence, which contains the positive emotion.

\begin{table*}
\centering
\begin{tabular}{p{15cm}} 
\toprule
\hline
YOFO~(\textit{ours})\\ 
\hline
\textbf{Aspect}: Location\\
\textbf{Label}:\textbf{~}Positive  \\   
\textbf{Pred}:\textbf{~}Positive \\
\textbf{Text}: i booked on lastminute . com for just £ 67 . 40 for 2 night double room . \textcolor{cyan}{very clean , \ul{good location}} , nice breakfast . if you flight to tegel airport 21 min by bus 109 to stop near hotel . from schonefeld bus and u - bahn 7about 50min . \textcolor{cyan}{close to potzdam} and lake wanasee . \\
\hline
\hline
\textbf{Aspect}: Service\\
\textbf{Label}:\textbf{~}Negative  \\   
\textbf{Pred}:\textbf{~}Negative \\
\textbf{Text}: very well located but very expensive for a 4 stars hotel . with a rack rate of 600 euros per night \ul{you can expect a }\textcolor{cyan}{\ul{minimum service}} . we had to bring our luggage up and down by ourselves and the internet broke down for half day . the amenities dodn ' t get refilled and the fresh towels were dirty . \textcolor{cyan}{\ul{the night staff was }}\ul{nice}\textcolor{cyan}{\ul{ although the day staff was very unpleasant}} . \\
\hline
\hline
\textbf{Aspect}: Cleanliness\\
\textbf{Label}:\textbf{~}Negative  \\   
\textbf{Pred}:\textbf{~}Negative \\
\textbf{Text}: i typically do n ' t expect a lot from airport hotels . these hotels typically have a very high volume of guests who typically do n ' t stay for much more than a couple of nights . in sum , i was n ' t expecting a lot by staying here . however , \ul{i was extremely }\textcolor{cyan}{\ul{disappointed}}\ul{ in }\textcolor{cyan}{\ul{the cleanliness of the room} ,} especially for such a reputable and high quality brand such as sheraton ( starwood ) . \textcolor{cyan}{\ul{the room itself i}}\ul{ found }\textcolor{cyan}{\ul{to be generally clean}}\ul{ , however }\textcolor{cyan}{\ul{the bathroom}}\ul{ was the }\textcolor{cyan}{\ul{worst} . there was a lot of loose hair in the bathroom , the mirror had water spots on it , and there were dirty towels hanging from the top of the shower} . i basically felt as if the room had not been cleaned by housekeeping staff before my check in . i also felt like the front desk staff was not very welcoming . i felt like i was a burden to the staff by checking in . the welcome was not warm or informative . nothing more occurred here other than an exchange of my id , credit card and a room key . not even an exchange of smiles . even though i will not recommend this hotel , i would hate to say , " do n ' t stay here " based on my single overnight experience . \\
\hline
\bottomrule
\end{tabular}
\caption{Case studies on Hotel Review dataset. Golden rationales are marked using \ul{underline}. Our rationales are highlighted using \textcolor{cyan}{cyan}.}
\label{table:decode_hotel}
\end{table*}

\subsection{The Conflict between ACC and F1 in the Hotel Review Dataset}
\label{sec:case_hotel}


In Section~\ref{sec:important}, we have already shown the effectiveness of extracted rationales in the BeerAdvocate dataset.
However, as we can see in Table~\ref{table:hotel}, our model achieves pretty low token-level f1 scores for rationale extraction.
How can a model with such low f1 scores, i.e. $\sim$50\%, perform well on downstream tasks?

We compare the test set between the aforementioned datasets and show the results in Table~\ref{table:hotel_statistic}.
We observed that not only is the label distribution in the Hotel Review dataset more balanced ($\sim$1:1 v.s. $\sim$41:1), but the input texts are also longer ($\sim$33 tokens more) than those in the BeerAdvocate dataset, potentially enhancing BERT's performance.

\begin{table}[htbp!]
\centering
\begin{tabular}{l|lll|lll}
\toprule
 & \multicolumn{3}{c|}{Beer}  & \multicolumn{3}{c}{Hotel} \\ \hline
 & Appearance & Aroma&	Palate&	Locations&	Services&	Cleanliness \\ \hline
\#avg tokens&	126.4&	126.9&	128.2&	154.8&	152.2&	147.2 \\
\#pos samples&	923&	848&	785&	104&	101&	97 \\ 
\#neg samples&	13&	29&	20&	96&	98&	99 \\ 
\bottomrule
\end{tabular}
\caption{The statistics on BeerAdvocate and Hotel Review test set.}
\label{table:hotel_statistic}
\end{table}

Furthermore, we decode the extracted rationales by YOFO in the Hotel Review dataset and show them in Table~\ref{table:decode_hotel}.
Based on our observations, the rationales generated by YOFO exhibit two issues compared to the human-annotated ones.
First of all, they are either excessively inclusive, including additional meaningful sentences, or incomplete.
For example, in the location aspect, the golden rationale is "good location".
However, YOFO selects not only the golden rationale but also other supporting phrases like "close to Potzdam", which could also be considered as a rationale.
In the service aspect, the golden rationales are transformed into sentences.
YOFO highlights the key phrase in the first golden sentence, namely "minimum service," and most of the second golden sentence except for the positive word "nice".
So it is evident that the rationale annotations in the dataset are inconsistent which causes the performance degradation.

Additionally, another observation demonstrates that there are other meaningful sentences that could serve as support for prediction, but they were not annotated by humans. For example, in the cleanliness aspect, the golden rationales are the concluding sentences. YOFO selects some concluding words and phrases from the golden rationale, as well as other specific behaviors, which align more closely with human reasoning.

\subsection{How to Decay ?}
\label{sec:decay_compare}
\subsubsection{Towards different token decay strategies.}
In this section, we will discuss different token decay strategies. 
Following the definition in Section \ref{section:LC}, we propose several decay methods in the RG part:

\paragraph{Linear}
The sparsity level decreases linearly from $1$ to $s$ in the RG part.
This can be expressed as $l_i=1 - (1-s)\times\frac{i-k}{d}$, if $k < i \le k+d$.

\paragraph{Exponential}
The sparsity level decreases exponentially from $1$ to $s$ in the RG part.
This can be formulated as $l_i=s^{\frac{i-k}{d}}$, if $k < i \le k+d$.

\paragraph{Logarithmic}
The sparsity level decreases logarithmically from $1$ to $s$ in the RG part.
This can be formulated as $l_i=\ln[\frac{\textit{e}^s(i-k)+\textit{e}(k+d-i)}{d}]$, if $k < i \le k+d$.

We select $\{0, 3, 6, 9, 12\}$ as our experimental setting pool, from which we randomly choose the start and end layers for the RG part.

Table~\ref{table:decay} demonstrates that Log decay tends to achieve higher accuracy compared to Linear and Exp decay under the same settings. 
When decaying from the embedding layer, Log decay is the optimal choice since its slow decay allows for more interactions between tokens.
Whereas, when decaying until the final layer, Exp decay is the most suitable choice.
Linear decay can be confidently applied when the RG part is located in the middle of the model. 
The optimal performance is obtained when the RG part spans from the third to the sixth layer with the Log decay.

In a word, the RG part achieves the best performance when it is situated in the middle of the model, specifically between the third and sixth layers using Log decay.

But some questions arise: Why is log decay the best when the token decays from the embedding layer? Why is exp decay the best when the token decays to the last layer?
Let's start with having a look at the learned token decay strategy from YOFO-final.

\begin{table*}[]
\centering
\begin{tabular}{l|l|llll|llll|llll}
\toprule
\multicolumn{1}{c|}{\multirow{2}{*}{from}} & \multicolumn{1}{c|}{\multirow{2}{*}{to}} & \multicolumn{4}{c|}{Linear}                                   & \multicolumn{4}{c|}{Exponential}                              & \multicolumn{4}{c}{Logarithmic}                               \\ \cline{3-14} 
\multicolumn{1}{c|}{}                      & \multicolumn{1}{c|}{}                    & Val ACC       & P             & R             & F1            & Val ACC       & P             & R             & F1            & Val ACC       & P             & R             & F1            \\ \hline
\multirow{4}{*}{0}                         & 3                                        & 87.5          & 71.4          & 55.7          & 62.6          & 88.2          & 65.9          & 48.8          & 56.1          & \textbf{88.3} & \textbf{73.9} & \textbf{57.3} & \textbf{64.6} \\
                                           & 6                                        & 87.6          & 77.0          & 57.8          & 66.0          & 88.1          & 67.4          & 51.1          & 58.1          & \textbf{88.3} & \textbf{84.3} & \textbf{65.2} & \textbf{73.6} \\
                                           & 9                                        & 88.4          & 74.8          & 57.7          & 65.1          & 88.1          & 64.7          & 49.6          & 56.2          & \textbf{88.7} & \textbf{79.7} & \textbf{63.4} & \textbf{70.6} \\
                                           & 12                                       & 88.1          & 63.6          & \textbf{60.0} & 61.7          & 87.8          & \textbf{67.2} & 58.2          & \textbf{62.4} & \textbf{89.3} & 59.7          & 54.1          & 56.8          \\ \hline
\multirow{3}{*}{3}                         & 6                                        & 88.5          & 89.7          & 69.8          & 78.5          & \textbf{89.0} & 90.1          & 70.6          & 79.4          & 88.0          & \textbf{91.3} & \textbf{73.0} & \textbf{81.1} \\
                                           & 9                                        & 88.3          & \textbf{91.7} & \textbf{70.3} & \textbf{79.6} & 87.6          & 87.5          & 67.4          & 76.1          & \textbf{89.5} & 85.9          & 68.2          & 76.1          \\
                                           & 12                                       & \textbf{89.4} & 63.2          & 54.5          & 58.5          & 89.1          & \textbf{70.0} & \textbf{60.0} & \textbf{64.6} & 89.2          & 61.4          & 49.2          & 54.6          \\ \hline
\multirow{2}{*}{6}                         & 9                                        & 88.7          & 88.0          & \textbf{69.4} & \textbf{77.6} & 88.6          & \textbf{89.0} & 64.5          & 74.8          & \textbf{89.1} & 87.7          & 66.2          & 75.5          \\
                                           & 12                                       & \textbf{89.8} & 72.5          & 53.5          & 61.5          & 89.4          & \textbf{81.9} & \textbf{67.3} & \textbf{73.9} & 89.7          & 56.1          & 44.7          & 49.8          \\ \hline
9                                          & 12                                       & \textbf{90.1} & 56.2          & 33.8          & 42.2          & 89.5          & \textbf{58.9} & \textbf{51.1} & \textbf{54.7} & 89.7          & 52.5          & 41.7          & 46.4          \\ \bottomrule
\end{tabular}
\caption{The results of different decay algorithms on decorrelated BeerAdvocate~\cite{beer} dataset Appearance aspect. "from" and "to" indicate the RG part starts and ends at this layer, respectively. {\textbf{Bold}} indicates the best results among different decay methods with the same start and end layer for the RG part.}
\label{table:decay}
\end{table*}

\subsubsection{The learned token decay strategy in YOFO-final}
\begin{figure}[htbp!]
	\centering
	\subfloat[Appearance aspect]{\includegraphics[width=.33\columnwidth]{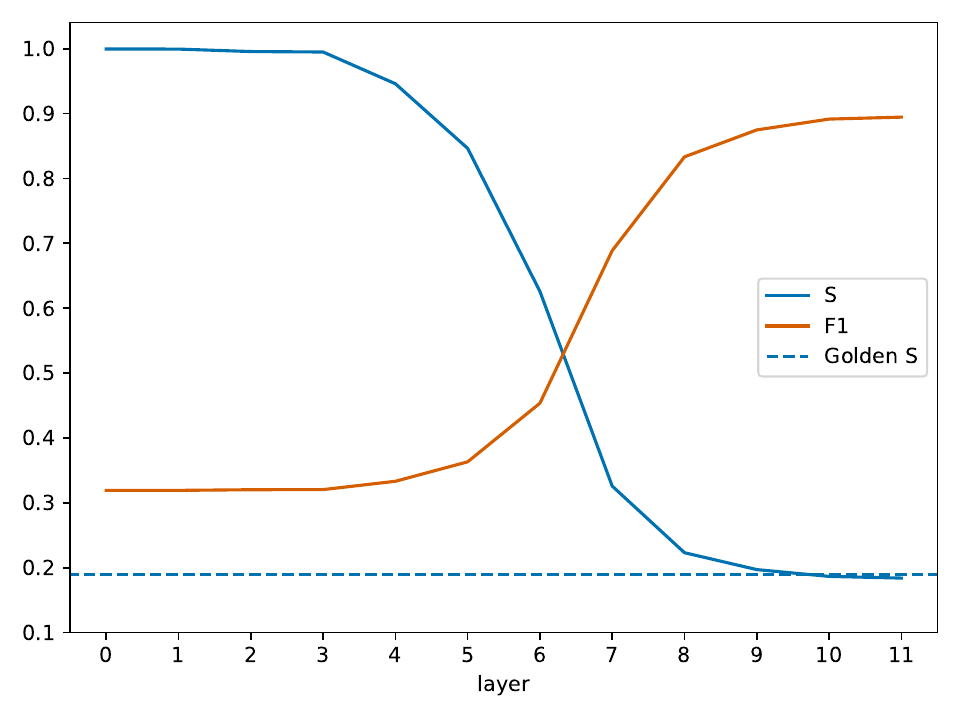}\label{fig:final_beer0}}
	\subfloat[Aroma aspect]{\includegraphics[width=.33\columnwidth]{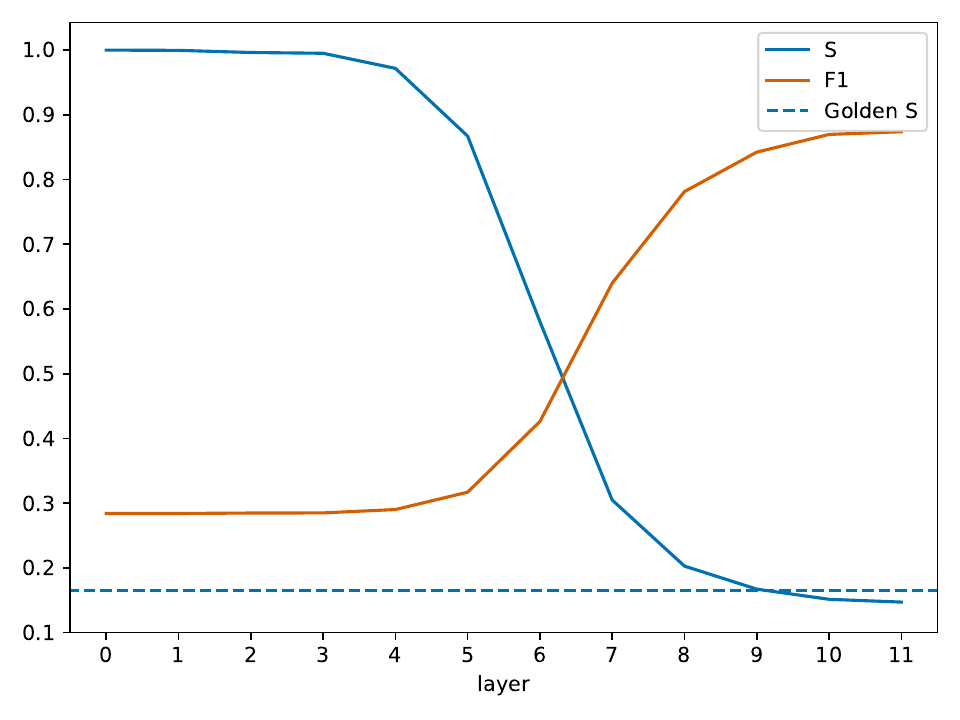}\label{fig:final_beer1}} 
	\subfloat[Palate aspect]{\includegraphics[width=.33\columnwidth]{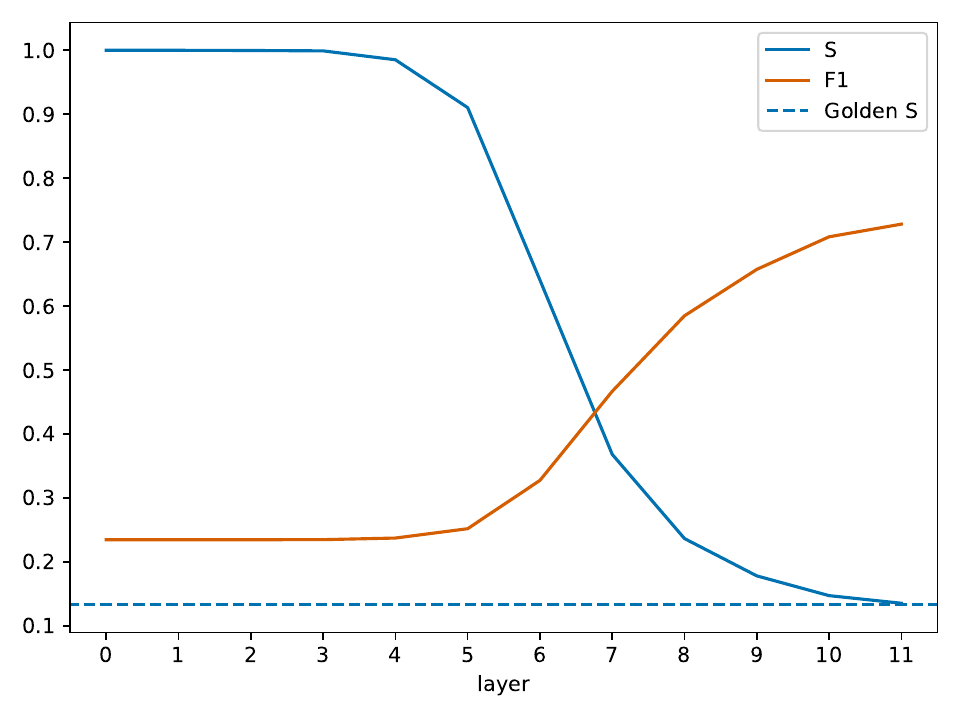}\label{fig:final_beer2}} 
	\caption{The learned length configuration of YOFO-final on BeerAdvocate dataset~\cite{beer}. Blue and orange lines denote the sparsity level and token-level f1 respectively. Green dash means the sparsity level of human-annotated rationales.}
 \label{fig:final_beer}
\end{figure}

We visualize the learned token decay strategy in YOFO-final in Figure~\ref{fig:final_beer}.
It can be observed that at the initial stage, with 100\% tokens selected, the sparsity level remains close to 1.0 while the f1 score is relatively low.
Throughout the rest of the model, the sparsity level experiences a slow decrease initially, followed by a rapid descent in the middle section, and finally slows down again to reach a certain level of sparsity in the last few layers.
This gradual decrease pattern does not strictly conform to any single decay mode discussed above.
Instead, it appears to be a combination of Logarithmic (for the initial descent layers), Linear (for the fastest descent segment), and Exponential (for the final few layers), which partly aligns with the discoveries in Section~\ref{sec:decay_compare}.

The question now becomes: Why do tokens decay slowly in the first few layers and the last few layers?

\subsubsection{What Does YOFO Learn in Each Layer ?}
\label{sec:case_layer}

\begin{table*}
\centering
\begin{tabular}{p{15cm}} 
\toprule
\hline
YOFO~(\textit{ours})\\ 
\hline
\textbf{Aspect}: Appearance\\
\textbf{Label}:\textbf{~}Positive  \\   
\textbf{Pred}:\textbf{~}Positive \\
\textbf{Text}: 10 / 3 / 2009 500ml bottled aug 31 / 09 . \ul{pours a hazy orange , and this thing is damn delicious .} incredibly easy to drink and so damn good . not a hint of the alc . \% at all . goes down so smooth . i loved the hints of grapefruit and bitterness to this beauty . worth the hype , and then some . \\ \hline

\hline

\textbf{Layer 0}: 10 / 3 / 2009 500ml bottled aug 31 / 09 hazy orange damn delicious damn beauty 
\\ \hline

\textbf{Layer 1}: 10 / 3 aug 31 / 09 a delicious easy loved 
\\ \hline

\textbf{Layer 2}: 10 pours a hazy orange delicious beauty 
\\ \hline

\textbf{Layer 3}: 10 a orange 
\\ \hline

\textbf{Layer 4}: 10 a 
\\ \hline

\textbf{Layer 5}: 10 a hazy orange
\\ \hline

\textbf{Layer 6}: 10 pours a hazy orange
\\ \hline

\textbf{Layer 7}: a hazy orange , and
\\ \hline

\textbf{Layer 8}: pours a hazy orange , and this damn delicious
\\ \hline

\textbf{Layer 9}: pours a hazy orange , and this thing is damn delicious
\\ \hline

\textbf{Layer 10}: 10 / 3 / 2009 500ml bottled aug 31 / 09 . a hazy
\\ \hline

\textbf{Layer 11}: 10 / 3 / hazy 
\\ \hline

\bottomrule
\end{tabular}
\caption{Case studies on Hotel Review dataset. Golden rationales are marked using \ul{underline}. Rationales generated by YOFO in layer k are shown after \textbf{Layer k}.}
\label{table:beer_layer}
\end{table*}

\begin{figure}[htbp!]
	\centering
	\subfloat[]{\includegraphics[width=.4\columnwidth]{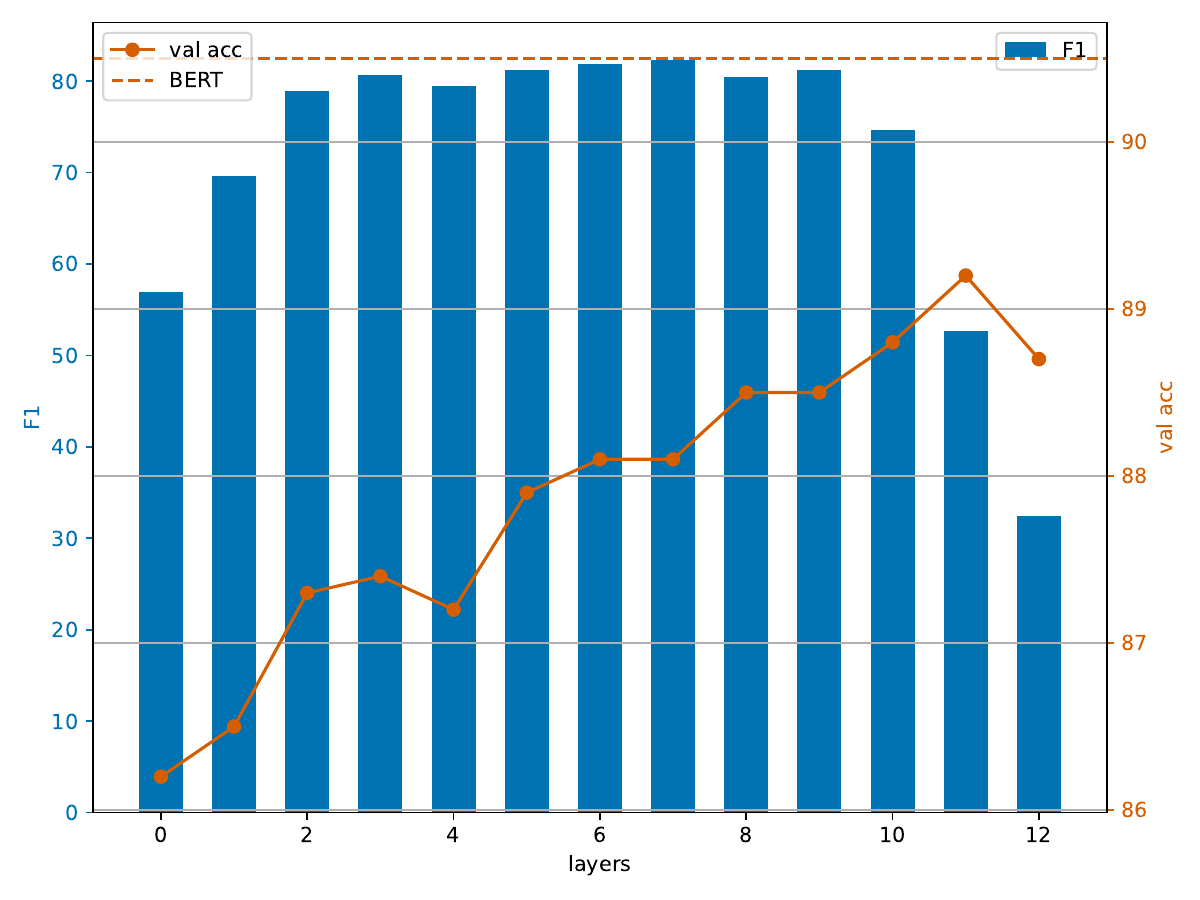}\label{beer_curve}}
	\subfloat[]{\includegraphics[width=.4\columnwidth]{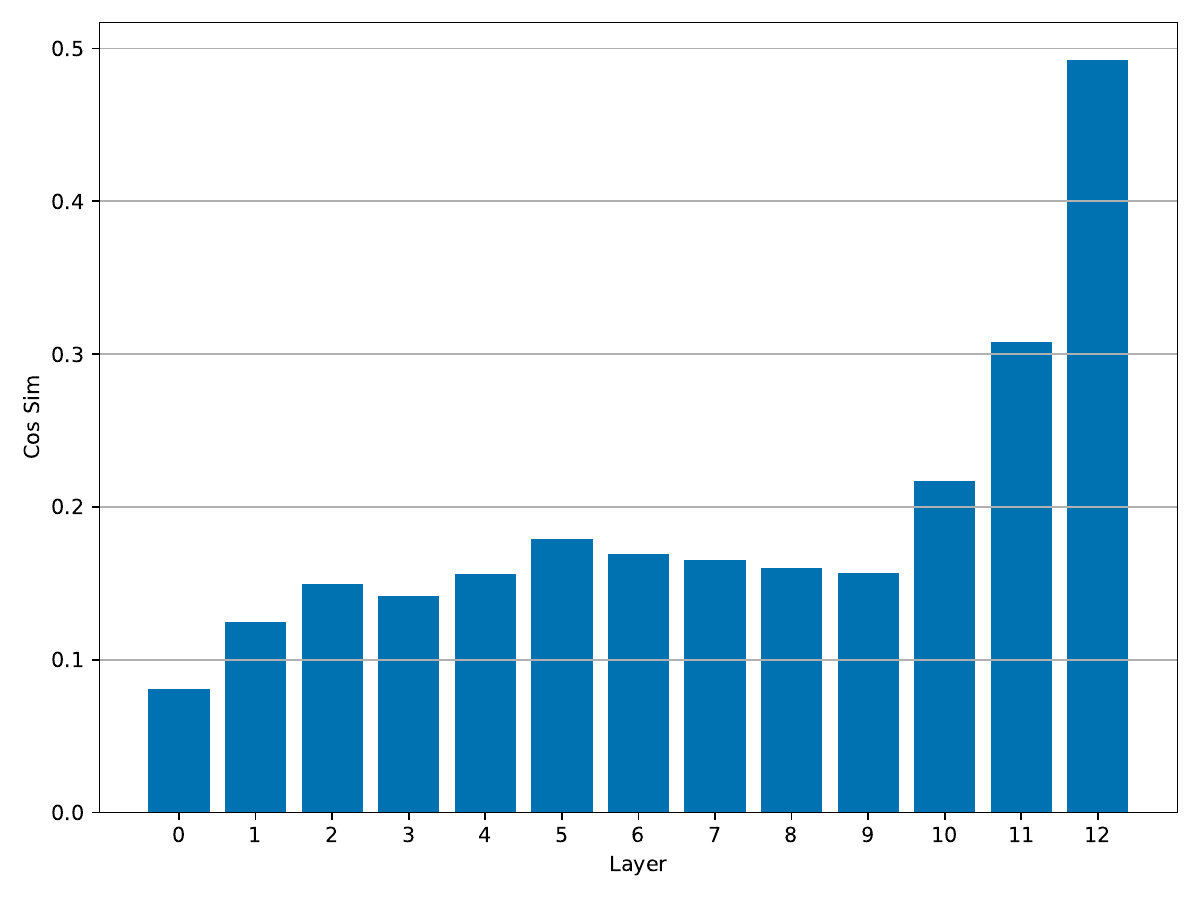}\label{fig:intra}} 
	\caption{(a). The performance curves on the decorrelated BeerAdvocate dataset~\cite{beer} Appearance aspect. "F1" denotes the token-level F1 scores for rationale extraction on the test set. "Val Acc" means the accuracy on the validation set. (b). Intra-layer similarity on decorrelated BeerAdvocate dataset~\cite{beer}.}
\end{figure}

In order to investigate what will happen if tokens decay in the first and last few layers, we deployed Cliff decay, which suddenly decreases sparsity to the target value, on pre-trained language models.
As defined in Section~\ref{section:LC}, the final sparsity $s$ was set to 0.13 with decay layer $k$ changing from 0 to 12. 
It is noteworthy that when $k=12$, sparsity control was not used, meaning that naive BERT was used. 
Conversely, when $k=0$, we anticipated achieving sparsity of $s$ after the word embedding layer, indicating no interactions would not occur between tokens in this setting.
We draw the performance curves in Figure~\ref{beer_curve} and decode the extracted rationales in Table~\ref{table:beer_layer}.

As shown in Figure~\ref{beer_curve}.
As the decay layer $k$ increases, validation accuracy improves.
The validation accuracy when $k=12$ is much lower than the original BERT model~(orange dash), which demonstrates the model suffers from overfitting.
Token-level F1 is exceptionally low when token deletion occurs at the beginning of several layers. 
We believe the reason is that tokens fail to fully interact with others in time.
However, token-level F1 drops abruptly when the deletion occurs in the last few layers. 
We believe this happens because the interactions between tokens are so strong that the model feels confused when deleting tokens.
We will further demonstrate the reason in the following part.
As such, deleting tokens at the middle part of pre-trained language models is better.


We also decode the rationales from each layer and present them in Table~\ref{table:beer_layer}.
From the table, we observe that during the initial layers, the model tends to select words or phrases that imply the sentiment of the label ("positive"). As we progress to the middle layers, the model gradually highlights the golden rationales by recognizing important tokens. By layers 8 and 9, the model can predict almost all of the ground truth. However, in the final two layers, the model becomes more confused about identifying important tokens compared to the previous layers.

To further investigate the failure of rationales from the last two layers, we train a BERT model on the BeerAdvocate dataset for the Appearance aspect. We then compute token similarity across different BERT layers using the corresponding dev set. This includes both intra-layer and inter-layer similarities. Intra-layer similarity measures the similarity of tokens' representations within the same layer, while inter-layer similarity compares the same token's representation between the $i$-th and $(i-1)$-th layers. 

As shown in Figure~\ref{fig:intra}, we observe a significant increase in terms of intra-layer token similarity in the last two layers. This high degree of similarity among tokens confuses the model as it struggles to determine the importance of each token. Notably, tokens in the final layer (Layer 11) exhibit the highest similarity, resulting in extremely poor rationale selection within this layer.

\section{Conclusion}
In this paper, we propose a framework called You Only Forward Once (YOFO) that builds upon the relaxed definition of rationale, considering it as support for predictions.
We also identify that the generate-then-predict paradigm experiences performance degradation due to both interlocking and spurious correlations.
To address the above problems, YOFO simultaneously generates rationales and predictions.
Our results on two unsupervised rationale extraction datasets demonstrate that our method achieves remarkable performance and outperforms most two-phase RNP-based models.
Additionally, experiments on the skewed and correlated BeerAdvocate Dataset show that our methods successfully mitigate interlocking and spurious correlations. 
Case studies further validate these findings. 
Through analysis of our framework, we discover that removing tokens too early or too late can be detrimental to performance, suggesting that removing tokens in the middle part with log decay is more effective.

\begin{acks}
This work was supported in part by the Natural Science Foundation of China (No.62006251) and the Natural Science Foundation of Hunan Province (No.2021JJ40783), the National Key Research and Development Program of China (No.2021YFF1201200),  the Science and Technology Major Project of Changsha (No.kh2202004). This work was carried out in part using computing resources at the High-Performance Computing Center of Central South University.
\end{acks}

\bibliographystyle{ACM-Reference-Format}
\bibliography{samples/reference}

\appendix

\section{Implementation Details}
\subsection{Main Experiments}
\label{sec:imp_main}
Our methods are implemented by Pytorch~\cite{pytorch} and huggingface transformers library~\cite{wolf2019huggingface}.
We use BERT~\cite{JacobDevlin2018BERTPO} as our pre-trained language model.
We use the AdamW~\cite{AdamW} optimizer with a learning rate $3e-5$ and weight decay $0.0$.
The Cliff decay is equipped in our model, in which the $k$ is set to $9$.
The hyperparameter controlling sparsity and contiguous level are picked from \{0.3,0.7,1,3,7,10,12,15\} by grid search.
The batch size is set to 64 and the maximum sequence length is 256.
Apart from the last two aspects of the Hotel Review dataset, we train the model for 10 epochs. For the last two aspects of the Hotel Review dataset, we only ran 5 epochs due to the large number of samples they contained.
The model in the final epoch is selected.

\subsection{Skewed Decorrelated BeerAdvocate Dataset}
\label{sec:imp_skew}
To verify whether models suffer from interlocking, previous researchers propose to give the predictor a bad initialization~\cite{interlocking}.
To achieve this, they use the first sentence along with the label to train the model for a few epochs and get the overfitted parameters in the predictor.
Then the normal training procedure is applied to the overfitted system.
the Skew\textit{k} means they pre-train the predictor \textit{k} epochs and then back to the normal training procedure.
Normally, \textit{k} is set to 10, 15, and 20, and batch size is set to 500.
However, We cannot set the batch size to 500 in our setting since the BERT is such a large model.
Alternatively, we pre-train our model in similar steps $x$ with previous works, where $x=\frac{\#samples}{500}\times\textit{k}$.
Therefore, in our setting, the model overfits 600/900/1200 steps, similar training steps as DR~\cite{DR}, and then the normal training procedure in Appendix~\ref{sec:imp_main} is executed.


\end{document}